\documentclass{article} 
\usepackage{iclr2027_conference,times}

\usepackage{amsmath,amsfonts,bm}

\def\eqref#1{equation~\ref{#1}}

\def\1{\bm{1}}

\DeclareMathAlphabet{\mathsfit}{\encodingdefault}{\sfdefault}{m}{sl}
\SetMathAlphabet{\mathsfit}{bold}{\encodingdefault}{\sfdefault}{bx}{n}

\usepackage{hyperref}
\usepackage{url}
\usepackage{booktabs}
\usepackage{array}
\usepackage{enumitem}
\usepackage{graphicx}
\usepackage{multirow}
\usepackage{algorithm}
\usepackage{algorithmic}

\newcolumntype{C}[1]{>{\centering\let\newline\\\arraybackslash\hspace{0pt}}m{#1}}
\newcolumntype{L}[1]{>{\raggedright\arraybackslash}m{#1}}

\title{Where to Learn and How Far to Roll Out: Efficient Adaptive Training for GUI Agents}

\author{
Yuxiang Chai\textsuperscript{12*} \quad
Quyu Kong\textsuperscript{1*} \quad
Panrong Tong\textsuperscript{1*} \quad
Steven Hoi\textsuperscript{1} \quad
Yue Wang\textsuperscript{1$\dagger$} \\
\textsuperscript{1} Alibaba Token Hub, Alibaba Group \quad
\textsuperscript{2} MMLab, The Chinese University of Hong Kong \\
\textsuperscript{*} Equal Contribution \quad
\textsuperscript{$\dagger$} Corresponding Author
}

\iclrpreprint

\renewcommand{\headrulewidth}{0pt}

\begin{document}

\maketitle

\begin{abstract}
Trajectory-level GRPO is a common approach for training multi-step GUI
agents, but repeatedly rolling out complete trajectories from the task root
can be inefficient: task-level rewards provide coarse supervision over which
intermediate decisions require improvement, while already-mastered prefixes
are repeatedly re-executed before reaching difficult states. We introduce a
two-stage adaptive training framework that jointly addresses
\emph{where to learn} and \emph{how far to roll out}. We first construct
reusable intermediate checkpoints from successful expert trajectories and
select informative pivots using \emph{action disagreement}, which measures
uncertainty directly from sampled GUI actions. Stage I performs one-step RL at
these pivots to improve uncertain local decisions. Stage II then progressively
expands policy responsibility through adaptive Local, Suffix, and Root
training according to task progress. Recoverable checkpoints enable repeated
training from intermediate states without replaying the preceding trajectory.
Across MobileGym and CUA-Gym-Hub, our method reduces environment interaction
by 56--61\% relative to Vanilla GRPO while improving task
performance, demonstrating that jointly allocating training across states and
rollout ranges can substantially improve GUI-agent RL efficiency.
\end{abstract}

\section{Introduction}

\begin{figure}[t]
    \centering
    \includegraphics[width=0.98\linewidth]{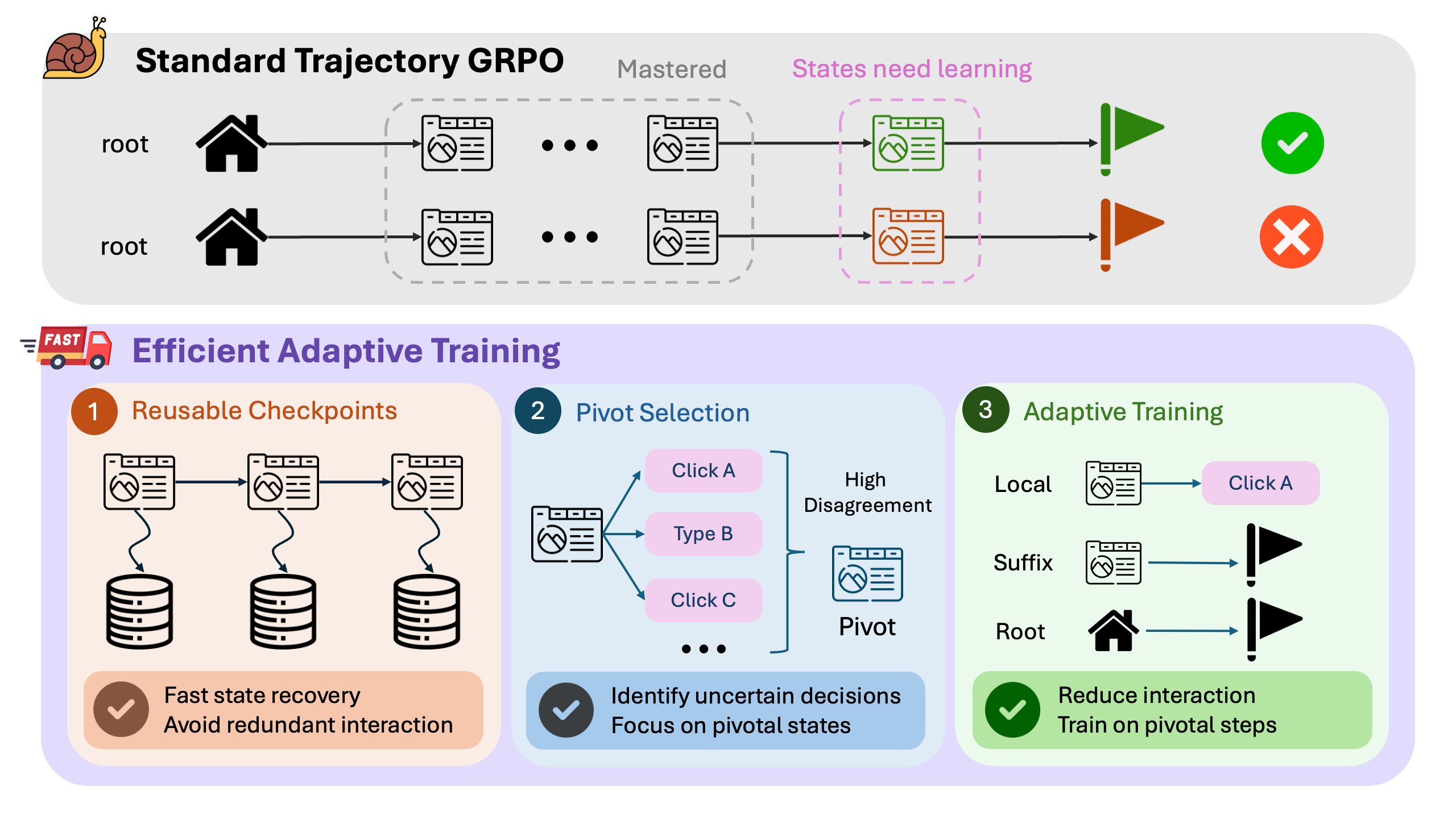}
    \caption{Overview of standard trajectory-level GRPO and our efficient adaptive training framework. Vanilla GRPO repeatedly traverses mastered prefixes and provides sparse trajectory-level supervision. Our method instead combines reusable checkpoints, pivot selection, and adaptive training to reduce redundant interaction and focus learning on pivotal steps.}
    \label{fig:overview}
\end{figure}

Reinforcement learning has become an effective paradigm for improving agents
on multi-step interactive tasks. A common training setup starts the agent from
the task root, rolls out a complete trajectory, and uses the final task outcome
to construct policy updates over the decisions made along that trajectory.
While simple and effective, this trajectory-level training can be inefficient
for GUI agents, where every rollout requires repeated interaction with an
external environment and successful completion depends on a sequence of
interdependent decisions. We identify two main sources of inefficiency.
First, task-level outcomes provide only coarse supervision over which
intermediate decisions actually require improvement, making credit assignment
difficult across a multi-step trajectory. Second, repeatedly starting from the
task root re-executes prefixes that the policy may already handle reliably
before reaching the states that still need learning. As illustrated in
Figure~\ref{fig:overview}, efficient GUI-agent training therefore requires
deciding both \emph{where to learn} and \emph{how far to roll out}.

Recent work has explored intermediate-state learning through local pivot
optimization, uncertainty-guided branching, and rollback-based correction
~\citep{pivotrl,bpo,pivoarl,src}. These approaches show that intermediate
states can provide a more targeted unit for agent improvement, but several
challenges remain in GUI environments. Weak policies may rarely reach useful
downstream states on their own, existing selection signals may overlook states
where the policy is uncertain but reward variation has not yet appeared, and
recovering an intermediate GUI state through action replay can be both costly
and unreliable under dynamic interfaces. More importantly, existing methods
typically decide either which intermediate states should receive training or
where additional rollouts should branch. We instead study these two choices
jointly: \emph{where optimization should begin} and \emph{how much of the
remaining trajectory the policy should be responsible for}.

To address these issues, we introduce a two-stage adaptive training framework
for GUI agents. We first construct reusable intermediate checkpoints from
successful expert trajectories, which provide reliable downstream states and
local next-action references even when the current policy cannot yet reach
those states consistently. In Stage I, we identify informative states using
\emph{action disagreement}, which measures uncertainty directly from the GUI
actions produced by repeated policy samples, and perform one-step RL at the
selected pivots. This concentrates early optimization on uncertain local
decisions without requiring successful completion of the remaining task. Stage II then progressively expands these local improvements into multi-step
task-solving ability through adaptive hybrid training. Tasks with low success
continue to receive local updates, partially mastered tasks are trained from
intermediate pivots using suffix rollouts whose starting frontier moves
progressively earlier as the policy improves, and reliably solved tasks use
only lightweight root-level rollouts. In this way, the framework adapts both
the training location and rollout extent to the policy's evolving capability.
Recoverable checkpoints support both stages by allowing repeated training and
branching from intermediate GUI states without replaying the full preceding
trajectory.

Our contributions are threefold:
\begin{itemize}[leftmargin=5mm]
\item We introduce an \emph{action-disagreement} criterion that selects
informative intermediate states by measuring uncertainty directly from sampled
GUI actions.

\item We propose a progress-aware two-stage training framework that jointly
adapts \emph{where a rollout starts} and \emph{how much of the task it covers},
progressing from local pivot updates to suffix and root-level training as the
policy improves.

\item We develop a checkpoint-supported training pipeline for efficient
intermediate-state learning without replaying full trajectory prefixes,
improving both training efficiency and task performance across GUI
environments.
\end{itemize}

\section{Related Work}

\subsection{Training GUI Agents with Multi-Turn Reinforcement Learning}

Early native GUI agents primarily rely on supervised learning from large-scale interaction traces. UI-TARS~\citep{uitars} unifies perception, reasoning, and action modeling over GUI trajectories, while ShowUI~\citep{showui} learns a vision-language-action policy from curated GUI interaction data. More recent work increasingly introduces reinforcement learning to improve multi-step GUI execution. UI-R1~\citep{uir1} and GUI-R1~\citep{guir1} apply GRPO-style reinforcement fine-tuning to GUI action learning; MobileGUI-RL~\citep{mobileguirl} develops online GUI RL with trajectory-aware advantages; and MobileRL~\citep{mobilerl} further adapts GRPO to task difficulty and online exploration. DART-GUI~\citep{dartgui} studies stabilized and efficient multi-turn GUI RL at larger scale, while CUA-Gym~\citep{cuagym} provides large-scale verifiable environments for training computer-use agents with outcome-based RL. Qwen-UI-Agent~\citep{qwenuiagent} further scales online RL to trajectories exceeding one hundred interaction steps across mobile, desktop, and web environments. These methods demonstrate the effectiveness of trajectory-level reinforcement learning for GUI agents. However, multi-turn training commonly relies on complete environment interaction trajectories and task-level outcomes, making both environment sampling and fine-grained credit assignment increasingly costly as task length grows. Our work studies a complementary question: rather than improving the underlying trajectory-level objective, how should training budget be allocated to the intermediate decisions that still require learning?

\subsection{Learning and Recovering Intermediate Agent States}

Recent work has explored intermediate states as a more targeted unit for agent learning. PivotRL~\citep{pivotrl} extracts candidate states from expert or SFT trajectories, selecting pivots according to local reward variance and applying one-step on-policy optimization. BPO~\citep{bpo} instead selects high-entropy decisions along policy-generated trajectories, snapshots the corresponding sandbox states, and branches multiple suffix rollouts from each state. PivoARL~\citep{pivoarl} uses self-reflection over failed trajectories to locate pivotal erroneous turns and retry from those decisions, while SRC~\citep{src} uses teacher review to identify the first harmful deviation of a speculative branch and performs rollback-based correction. Tree-GRPO~\citep{treegrpo} randomly branches from intermediate interaction steps to share common prefixes and obtain finer-grained training signals. Intermediate-state methods also depend on how the underlying environment is recovered. BPO directly exploits snapshot-and-restore support in executable sandboxes, whereas other approaches rely on resettable environments or action-prefix replay. In GUI research, OSExpert~\citep{osexpert} restores previous exploration nodes by replaying their stored action sequences. Although replay requires little environment-specific checkpointing support, it repeatedly incurs prefix interaction cost and can be fragile under dynamic GUI behavior. These methods primarily decide either which intermediate states to optimize
or where to branch rollouts. In contrast, our focus is to jointly
adapt the training location and the rollout extent as policy capability
changes.

\section{Efficient Adaptive Training}

Trajectory-level GRPO repeatedly rolls out complete trajectories from the task
root, even when only a subset of intermediate decisions still require
improvement. We therefore organize training around two questions:
\emph{where to learn} and \emph{how far to roll out}. Our framework addresses
both by reusing recoverable intermediate states, selecting informative pivots
through action disagreement, and progressively expanding training from local
updates to longer-horizon execution.

\begin{figure*}[t]
    \centering
    \includegraphics[width=\textwidth]{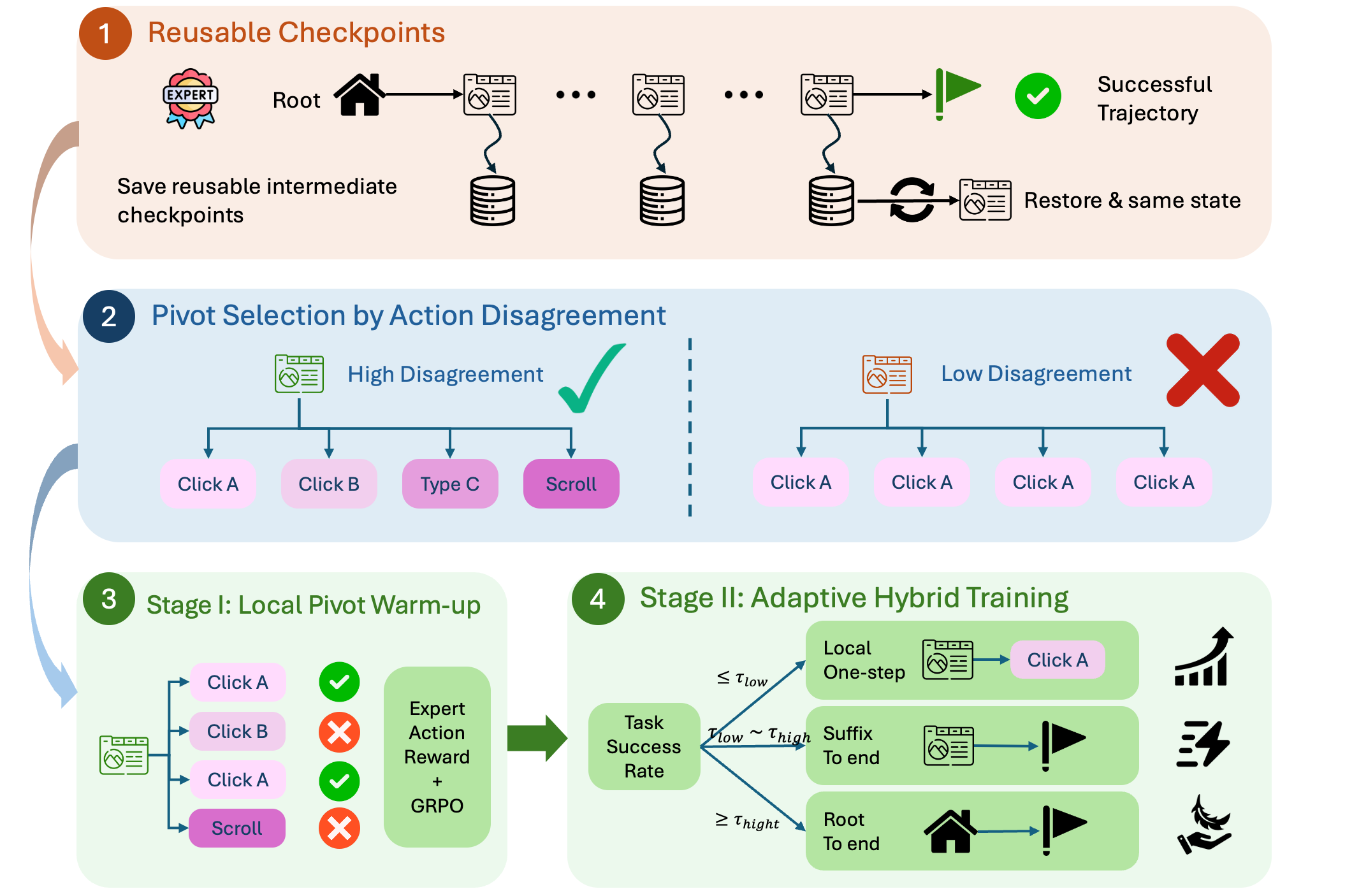}
    \caption{Training pipeline of our method. We construct reusable checkpoints from successful expert trajectories, select uncertain intermediate states using action disagreement, and perform a two-stage training procedure consisting of one-step local warm-up followed by adaptive hybrid training.}
    \label{fig:pipeline}
\end{figure*}

\subsection{Reusable Intermediate States}

Training from an intermediate GUI state requires recovering both the
policy-visible context and the underlying executable environment. Restarting
from the task root and replaying the preceding action sequence
incurs the same prefix cost, and may drift in dynamic interfaces due to
asynchronous loading, animations, or timing-dependent behavior. We therefore
save reusable checkpoints along successful trajectories. For a trajectory
$\tau=(o_0,a_0,o_1,a_1,\ldots,o_T)$, we define the checkpoint at step $t$ as
$c_t=(x,h_t,o_t,e_t)$, where $x$ denotes the task instance, $h_t$ the
policy-visible interaction history, $o_t$ the current observation, and $e_t$
a recoverable reference to the corresponding executable environment state.

Once created, a checkpoint can be restored repeatedly to generate independent
rollout branches without re-executing the preceding trajectory, amortizing the
cost of reaching deep intermediate states. As shown in
Table~\ref{tab:checkpoint-replay}, checkpoint restoration reduces the measured
recovery latency of a 10-step prefix by $81.1\%$ on MobileGym and $71.6\%$ on
CUA-Gym-Hub. Environment-specific checkpointing and validation details are
provided in the appendix.

\begin{table*}[t]
\caption{Checkpoint restoration latency compared with replaying a 10-step trajectory prefix.}
\label{tab:checkpoint-replay}
\centering
\small
\begin{tabular}{L{3cm}C{3cm}C{3cm}C{3cm}}
\toprule
\textbf{Environment} & \textbf{Checkpoint Restore} & \textbf{10-Step Replay} & \textbf{Time Reduction} \\
\midrule
MobileGym    & 2.97 s & 15.7 s & 81.1\% \\
CUA-Gym-Hub & 3.21 s & 11.3 s & 71.6\% \\
\bottomrule
\end{tabular}
\end{table*}

\subsection{Where to Learn: Selecting Pivots by Action Disagreement}
\label{sec:pivot-selection}

Similar to PivotRL~\citep{pivotrl}, we construct candidate training states from successful expert trajectories, which provide reliable next-action references and expose downstream states that the initial policy may rarely reach.

We select among these states based on the uncertainty of the current policy's GUI decision. This follows the uncertainty-guided intuition of BPO~\citep{bpo}, but instead of using first-token policy entropy, we measure uncertainty directly from the GUI actions produced by policy samples. This also differs from PivotRL's reward-variance criterion: a state may produce diverse actions while all current samples receive the same reward, such as $0/8$. In this case, the current group provides no reward variation, but the diverse actions indicate that the model is still uncertain about what to do at this state. As training changes the policy distribution, later iterations may produce mixed outcomes and thus useful learning signals from the same state.

Given $N$ policy samples at state $c_t$, let $n_j$ denote the number of
samples producing the $j$-th distinct GUI action. We define the empirical
action disagreement as

$$
\hat{D}(c_t)
=
1-
\frac{\sum_{j} n_j(n_j-1)}
{N(N-1)},
$$

which measures the probability that two sampled responses produce different
GUI actions. We select states with $\hat{D}(c_t)>\tau_{\mathrm{dis}}$ as \emph{pivots}. This directly captures uncertainty
in the policy's GUI actions without requiring reward variation to have already
appeared. The selected pivots are subsequently used for local updates and
suffix rollouts.

\subsection{How Far to Roll Out: Adaptive Hybrid Training}
\label{sec:adaptive-hybrid}

After selecting where to train, we determine how much of the remaining task
should be optimized from that state. Our routing follows two principles:
use the shortest rollout that still provides a useful learning signal, and
expand the policy's responsibility only after the current segment becomes
reliable. This yields a progression from local one-step updates, to suffix
rollouts from intermediate pivots, and finally to lightweight root-level
training.

For task $x$, let $\hat{p}(x)$ denote its root-level success rate:
$$
\hat{p}(x)
=
\frac{\#\text{successful rollouts}}
{\#\text{rollouts}}.
$$
Given thresholds $\tau_{\mathrm{low}}$ and $\tau_{\mathrm{high}}$, we use
$\hat{p}(x)$ to initialize the training regime:
$$
R(x)=
\begin{cases}
\text{Local}, & \hat{p}(x)<\tau_{\mathrm{low}},\\
\text{Suffix}, & \tau_{\mathrm{low}}\leq \hat{p}(x)\leq\tau_{\mathrm{high}},\\
\text{Root}, & \hat{p}(x)>\tau_{\mathrm{high}}.
\end{cases}
$$
This routing determines the initial rollout extent. Training then follows a
progressive schedule in which Local tasks may advance to Suffix or Root,
Suffix tasks move their rollout frontier toward earlier pivots, and tasks that
reach Root remain in the Root regime.

\paragraph{Local training.}
When root-level success is low, long rollout groups often collapse to the same
failure outcome and provide weak relative learning signal despite substantial
interaction cost. We therefore perform one-step RL at the selected pivots,
using the expert next action to construct an immediate local reward. After
local updates, the task is re-evaluated from the root and may remain in Local
training or advance to Suffix or Root as its task success improves.

\paragraph{Suffix training.}
Once the policy has acquired partial task-solving ability, longer rollouts
become more informative. We order the selected pivots along the expert
trajectory and begin from the deepest pivot, avoiding the already-reliable
prefix. From the current pivot, we generate suffix rollouts to task
termination. If the suffix success rate exceeds $\tau_{\mathrm{suffix}}$, the
rollout frontier moves one pivot earlier in the next round; otherwise, it
remains at the current pivot. Repeating this process progressively increases
the portion of the trajectory completed by the policy until training reaches
the earliest pivot and starts root rollout.

\paragraph{Root training.}
When the policy can reliably solve the task end-to-end, we use only a small
number of root-level rollouts to probe whether additional optimization remains
useful. If the rollout group exhibits outcome variation, it provides a valid
training signal and is used for policy updates. If all probes succeed
consistently, no update is performed. Once a task enters Root training, it
remains in this regime.

\subsection{Two-Stage Training}
\label{sec:two-stage}

Although Adaptive Hybrid Training can be applied directly to the base policy,
we find it beneficial to first improve uncertain local decisions before
introducing longer-horizon optimization. Our final method therefore consists
of two stages.

\paragraph{Stage I: Local pivot warm-up.}
We use the Action-Disagreement pivots selected in Section~\ref{sec:pivot-selection} and perform only
the one-step local updates described in Section~\ref{sec:adaptive-hybrid}. This stage concentrates
optimization on uncertain GUI decisions without requiring the policy to
complete the remaining task, providing a stronger initialization for
subsequent multi-step training.

\paragraph{Stage II: Adaptive hybrid training.}
Starting from the Stage-I policy, we apply the Adaptive Hybrid procedure from
Section~\ref{sec:adaptive-hybrid}. An initial set of root-level rollouts assigns each task to Local,
Suffix, or Root training. Local tasks are periodically re-evaluated from the
root, Suffix training progressively moves the rollout frontier toward earlier
pivots as the current suffix becomes reliable, and tasks that reach Root remain
in Root-level training.

\section{Experiments}

\subsection{Experimental Setup}
\label{sec:exp-setup}

\paragraph{Environments and task splits.}
We evaluate our method on two GUI environments, MobileGym~\citep{mobilegym} and CUA-Gym-Hub~\citep{cuagym}. For MobileGym, we construct our split from the official 256-task test set because the original training and test sets exhibit a substantial distribution gap. We divide these 256 tasks into two disjoint subsets of 128 tasks. After verifying task executability with the expert, 83 tasks from the first subset are retained for training, while the remaining 128 tasks are used exclusively for held-out evaluation. For CUA-Gym-Hub, we select 80 tasks from the official task pool for training and a disjoint set of 60 tasks for evaluation. In both environments, evaluation tasks are never used during RL training.

\paragraph{Policy and expert trajectories.}
All experiments start from a Qwen3.5-4B-based policy initialized from supervised fine-tuning. Importantly, the SFT data are collected from other environments and tasks and contain no trajectories from either MobileGym or CUA-Gym-Hub, ensuring that improvements on the two benchmarks come from the RL training studied in this work. To construct successful expert trajectories and intermediate training states, we use Qwen3.8-Max as the expert for MobileGym and GPT6-Astra for CUA-Gym-Hub. Only successfully completed and verified expert trajectories are used to construct reusable checkpoints and intermediate states.

\paragraph{Evaluation protocol.}
We evaluate policies from the task root without providing expert prefixes or intermediate checkpoints at test time. For each task, we perform four independent evaluation rollouts and report both \emph{Avg@4}, the average success rate across the four attempts, and \emph{Pass@4}, the fraction of tasks solved at least once within four attempts. We report results separately on the training-task set (Train) and the held-out evaluation set (Held-Out), and also measure training efficiency using both the number of training optimization samples and the number of environment interaction steps.

\paragraph{Training details.}
For pivot selection, we sample $N=8$ responses at each candidate state.
Reward Variance retains states with non-zero local reward variance, whereas Action Disagreement uses $\tau_{\mathrm{dis}}=0.45$, filtering weak action dispersion while preserving
states with substantial decision uncertainty.
Hybrid training is initialized using four root rollouts per task, with
$\tau_{\mathrm{low}}=0.25$ and $\tau_{\mathrm{high}}=0.75$ to assign tasks
to Local, Suffix, or Root regimes. Local tasks are re-evaluated from the root, while Suffix and Root tasks follow the progression defined in Section~\ref{sec:adaptive-hybrid}. Local tasks are re-evaluated with four root rollouts; Suffix training
uses eight rollouts and moves the frontier one pivot earlier when the previous
suffix success rate exceeds $\tau_{\mathrm{suffix}}=0.75$; Root training uses
only two lightweight root rollouts. For the two-stage method, Stage I performs one-step RL on Action-Disagreement pivots, and the checkpoint with the highest Train Avg@4 is used to initialize Stage II. We define \emph{training samples} as policy samples that enter optimization and \emph{environment steps} as GUI interaction steps executed during rollout generation.

\subsection{Where to Learn: Evaluating Pivot Selection}
\label{sec:pivot-selection-exp}

\begin{figure}[t]
    \centering
    \includegraphics[width=0.98\linewidth]{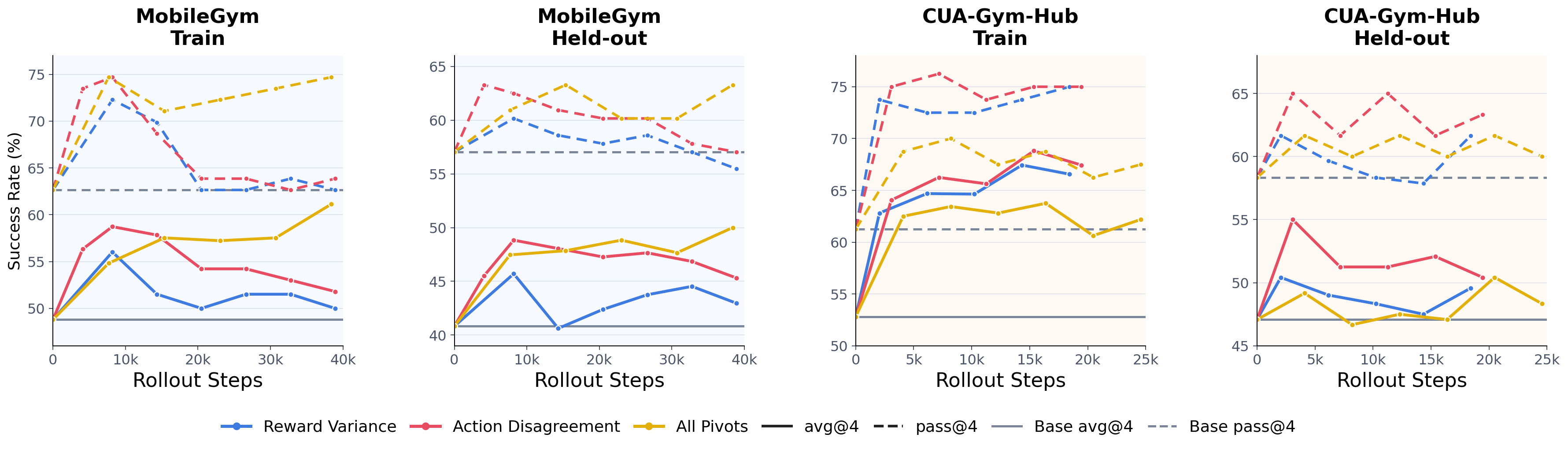}
    \caption{One-step pivot training with different state-selection strategies on MobileGym and CUA-Gym-Hub. All methods use the same local RL procedure and differ only in the selected intermediate states. Solid and dashed curves denote Avg@4 and Pass@4, respectively; gray lines show the common base performance.}
    \label{fig:pivot-selection}
\end{figure}

We first isolate the effect of pivot selection using one-step local RL. All
methods use the same expert trajectories, local reward, and GRPO training
procedure; they differ only in which intermediate states are selected for
training. We compare three strategies: \emph{All Pivots}, which trains on all
available intermediate states; \emph{Reward Variance}, which retains states
whose sampled actions produce non-zero local reward variance; and
\emph{Action Disagreement}, our method, which selects states where policy
samples produce diverse GUI actions. Figure~\ref{fig:pivot-selection} shows the full learning curves as a function
of cumulative rollout steps on MobileGym and CUA-Gym-Hub, while
Table~\ref{tab:pivot-selection} summarizes the checkpoint with the highest
Train Avg@4 for each method, together with its sample cost and corresponding
Train/Held-Out performance. Since all methods use the same one-step training objective,
differences primarily reflect the effectiveness of the state-selection
strategy rather than longer-horizon exploration.

\begin{table*}[t]
\caption{
Comparison of pivot-selection strategies on MobileGym and CUA-Gym-Hub.
Each metric cell reports Avg@4 / Pass@4 at the checkpoint with the highest Train Avg@4.
}
\label{tab:pivot-selection}
\centering
\small
\renewcommand{\arraystretch}{1.2}
\resizebox{\textwidth}{!}{
\begin{tabular}{l|ccc|ccc}
\toprule
\multirow{2}{*}{\textbf{Method}}
& \multicolumn{3}{c|}{\textbf{MobileGym}}
& \multicolumn{3}{c}{\textbf{CUA-Gym-Hub}} \\
\cmidrule(lr){2-4}
\cmidrule(lr){5-7}
& \textbf{Samples}
& \textbf{Train}
& \textbf{Held-Out}
& \textbf{Samples}
& \textbf{Train}
& \textbf{Held-Out} \\
\midrule

All Pivots
& 38.4K
& 61.14 / 74.70
& \textbf{50.00 / 63.28}
& 16.4K
& 63.75 / 68.75
& 47.08 / 60.00 \\

Reward Variance
& 8.2K
& 56.02 / 72.29
& 45.70 / 60.16
& 14.3K
& 67.44 / 73.75
& 47.50 / 57.87 \\
\midrule
\textbf{Action Disagreement (Ours)}
& 8.2K
& 58.73 / 74.70
& 48.83 / 62.50
& 15.4K
& \textbf{68.81 / 75.00}
& \textbf{52.08 / 61.67} \\

\bottomrule
\end{tabular}
}
\end{table*}

Across both benchmarks, Action Disagreement provides a stronger overall
accuracy-efficiency trade-off than Reward Variance on both Train and Held-Out tasks.
This supports our motivation that policy action uncertainty can identify
useful training states beyond those that already exhibit reward variation.
On MobileGym, All Pivots eventually obtains the highest Held-Out Avg@4, reaching
50.00\% after 38.4k rollout samples. Action Disagreement reaches a comparable
48.83\% with only 8.2k samples, approximately 21\% of the rollout budget,
while also outperforming Reward Variance at the same 8.2k-sample budget.

The advantage is more pronounced on CUA-Gym-Hub. Action Disagreement achieves
the highest Held-Out performance, reaching 55.00\% Avg@4 and 65.00\% Pass@4 with
3.1k rollout samples, while both Reward Variance and All Pivots peak at
50.42\% Held-Out Avg@4. Together, the learning curves and best-checkpoint summary
show that Action Disagreement provides a strong balance between training cost and downstream performance. We therefore use
Action-Disagreement pivots in the subsequent full-training experiments.

\begin{figure*}[t]
    \centering
    \includegraphics[width=0.99\textwidth]{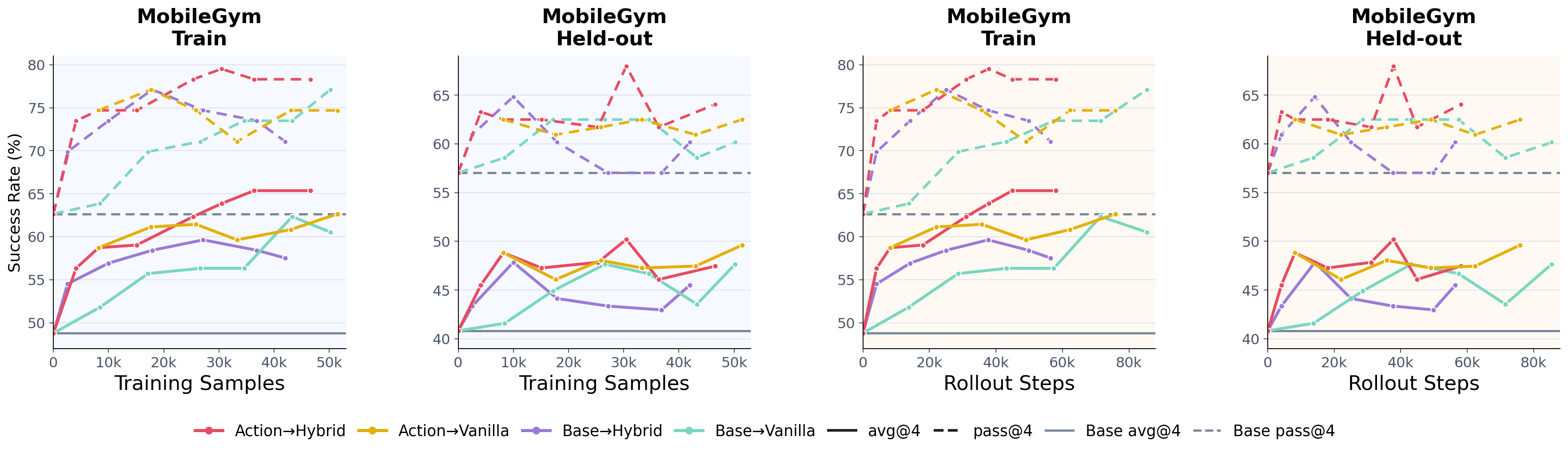}
    \vspace{1mm}

    \includegraphics[width=0.99\textwidth]{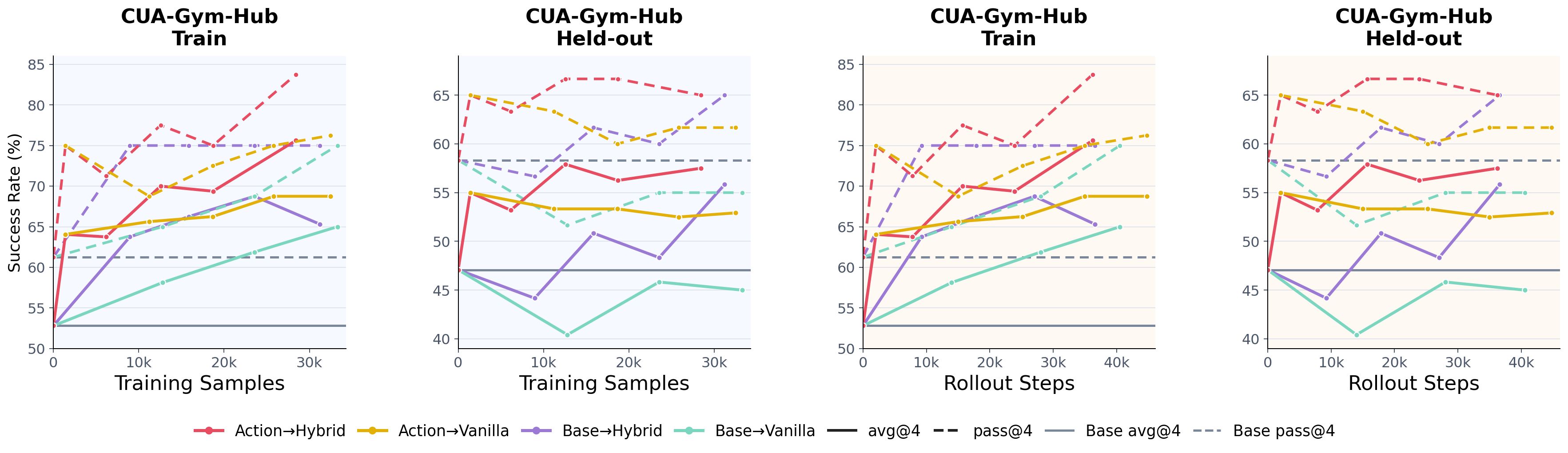}
    \caption{
    Full-training effectiveness and efficiency on MobileGym (top) and CUA-Gym-Hub (bottom). For each benchmark, the left two panels report Train and Held-out performance against cumulative training samples, while the right two report performance against cumulative rollout steps. Solid and dashed curves denote Avg@4 and Pass@4, respectively, and horizontal gray lines indicate base-policy performance.
    }
    \label{fig:full-training}
\end{figure*}

\subsection{Full Training: Effectiveness and Efficiency}

We next evaluate the complete training procedure in terms of both task
performance and training efficiency. We compare four configurations:
Base$\rightarrow$Vanilla, which applies standard trajectory-level GRPO
directly to the base policy; Base$\rightarrow$Hybrid, which applies Adaptive
Hybrid Training without Stage-I warm-up; Action$\rightarrow$Vanilla, which
performs Action-Disagreement one-step training followed by Vanilla GRPO; and
Action$\rightarrow$Hybrid, our full two-stage method. Figure~\ref{fig:full-training}
reports performance against cumulative training samples and environment rollout
steps on MobileGym and CUA-Gym-Hub.

\paragraph{Adaptive Hybrid vs.\ Vanilla GRPO.}
We first isolate the effect of adaptive rollout allocation by comparing
Base$\rightarrow$Hybrid with Base$\rightarrow$Vanilla. Across both benchmarks,
Hybrid reaches comparable or better task performance with substantially less
training and environment interaction. On MobileGym, Base$\rightarrow$Hybrid
reaches 58.43 Train Avg@4 using 17.9k training samples and 25.0k rollout steps,
whereas Base$\rightarrow$Vanilla requires 43.3k samples and 71.5k steps to
surpass the same level, reducing the two costs by 58.6\% and 65.0\%,
respectively. A similar trend appears on CUA-Gym-Hub, where
Base$\rightarrow$Hybrid reaches 63.75 Train Avg@4 with 9.0k samples and 9.2k
rollout steps, compared with 33.3k samples and 40.5k steps for Vanilla.

The same efficiency advantage is also reflected on held-out tasks. On
MobileGym, Base$\rightarrow$Hybrid reaches 47.85 Held-out Avg@4 with 10.0k
samples and 14.2k rollout steps, slightly above the 47.66 obtained by Vanilla
after 26.6k samples and 43.2k steps. On CUA-Gym-Hub,
Base$\rightarrow$Hybrid reaches 55.83 Held-out Avg@4, whereas the best
Base$\rightarrow$Vanilla checkpoint reaches 45.83. These results indicate that
adaptive rollout allocation reduces the cost of trajectory-level RL without
sacrificing downstream task performance.

\paragraph{Two-Stage Training.}
We next examine whether the Action-Disagreement warm-up further improves
multi-step training. Action$\rightarrow$Hybrid consistently reaches strong
performance earlier than Action$\rightarrow$Vanilla. On MobileGym, the full
method reaches 63.86 Train Avg@4 with 30.5k training samples and 37.8k rollout
steps, exceeding the best Action$\rightarrow$Vanilla result of 62.65 while
using 40.8\% fewer training samples and 50.2\% fewer environment interactions.
On CUA-Gym-Hub, Action$\rightarrow$Hybrid reaches 70.00 Train Avg@4 with
12.6k samples and 15.7k rollout steps, while Action$\rightarrow$Vanilla peaks
at 68.75 despite substantially larger training cost.

The benefit also transfers to held-out evaluation. On MobileGym,
Action$\rightarrow$Hybrid reaches 50.20 Held-out Avg@4, compared with 49.61
for Action$\rightarrow$Vanilla. On CUA-Gym-Hub, the full method reaches 57.92
Held-out Avg@4 after 12.6k samples and 15.7k rollout steps, whereas
Action$\rightarrow$Vanilla reaches 53.33 at a similar interaction budget and
does not improve beyond this level with additional training. Together, these
results suggest that the two stages are complementary: Stage I improves
uncertain local decisions, while Stage II expands these improvements into
multi-step task execution.

\begin{figure}
    \centering
    \includegraphics[width=0.99\linewidth]{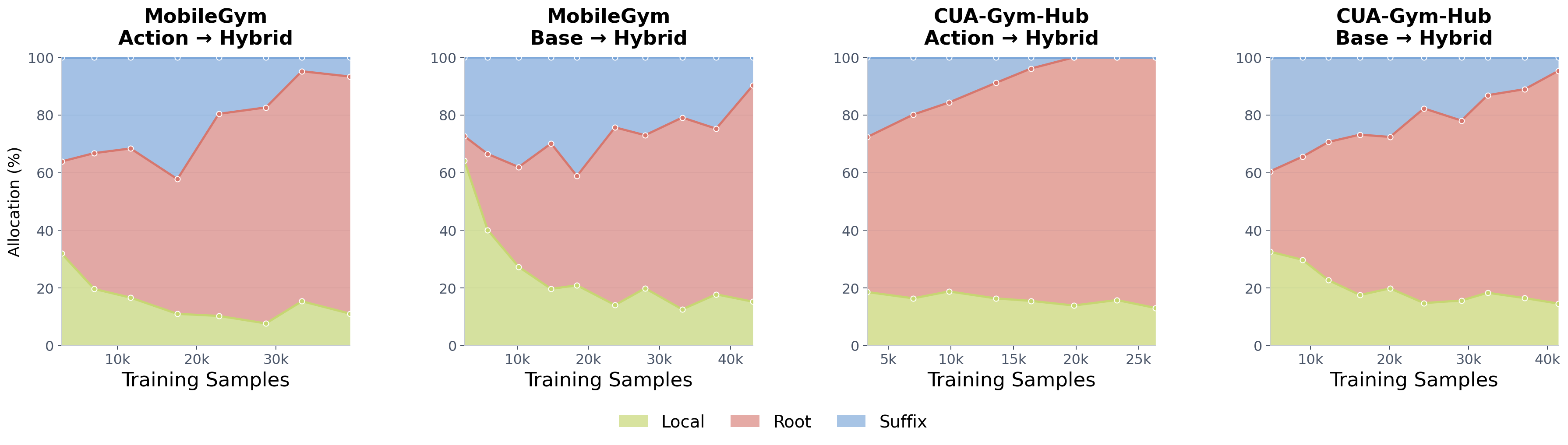}
    \caption{Evolution of training allocation in adaptive hybrid training on MobileGym and CUA-Gym-Hub. The four panels show how the proportions of Local, Root, and Suffix training change over cumulative training samples for Action$\rightarrow$Hybrid and Base$\rightarrow$Hybrid.
}
    \label{fig:allocation}
\end{figure}

\paragraph{Comparison with SFT.}
Because our framework also relies on successful expert trajectories, we compare
against supervised fine-tuning on the same expert data. The SFT baseline uses
all expert trajectory steps rather than only selected pivots. As shown in
Table~\ref{tab:full-training-summary}, SFT substantially improves Train performance on
both benchmarks, but its gains transfer less consistently to held-out tasks.
On MobileGym, SFT reaches 48.24 Held-out Avg@4, close to the 50.20 achieved by
our method. On CUA-Gym-Hub, both methods reach 70.00 Train Avg@4, yet
Action$\rightarrow$Hybrid achieves 57.92 Held-out Avg@4 compared with 50.83
for SFT, with Pass@4 increasing from 58.33 to 66.67. This suggests that the RL
procedure provides benefits beyond supervised imitation of the same expert
trajectories, particularly on held-out CUA-Gym-Hub tasks.
\vspace{-0.5em}
\paragraph{Adaptive Training Allocation.}
Figure~\ref{fig:allocation} shows how the training budget shifts across Local,
Suffix, and Root regimes as training progresses. For
Action$\rightarrow$Hybrid, the fraction of Root training increases from
31.96\% to 82.29\% on MobileGym and from 53.78\% to 86.97\% on CUA-Gym-Hub,
while Suffix training decreases from 36.12\% to 6.65\% and from 27.66\% to
0\%, respectively. This behavior is consistent with the intended progression
of Adaptive Hybrid Training: optimization begins with local or partial
trajectories and gradually shifts toward full-task execution as the policy
improves. Moreover, Action$\rightarrow$Hybrid begins Stage II with a larger
fraction of Root training than Base$\rightarrow$Hybrid on both benchmarks,
consistent with the Stage-I warm-up allowing more tasks to enter
longer-horizon training earlier.

\begin{table}[t]
\centering
\small
\caption{
Summary of full-training performance on MobileGym and CUA-Gym-Hub.
For each method, we select the checkpoint with the highest Train Avg@4 and
report its corresponding Train and Held-Out performance.
Each cell reports Avg@4 / Pass@4.
}
\label{tab:full-training-summary}
\begin{tabular}{L{3cm}|C{2.2cm}C{2.2cm}|C{2.2cm}C{2.2cm}}
\toprule
\multirow{2}{*}{\textbf{Method}}
& \multicolumn{2}{c|}{\textbf{MobileGym}}
& \multicolumn{2}{c}{\textbf{CUA-Gym-Hub}} \\
\cmidrule(lr){2-3}\cmidrule(lr){4-5}
 & Train & Held-Out & Train & Held-Out \\
\midrule
Base
& 48.80 / 62.65
& 40.82 / 57.03
& 52.81 / 61.25
& 47.08 / 58.33 \\

SFT
& \textbf{81.93 / 90.36}
& 48.24 / 65.63
& 70.00 / 73.75
& 50.83 / 58.33 \\

Base$\rightarrow$Vanilla
& 56.33 / 71.08
& 47.66 / 62.50
& 61.88 / 68.75
& 45.83 / 55.00 \\

Base$\rightarrow$Hybrid
& 56.93 / 73.49
& 47.85 / 64.84
& 65.31 / 75.00
& 55.83 / 65.00 \\

Action$\rightarrow$Vanilla
& 62.65 / 74.69
& 49.61 / 62.50
& 65.63 / 68.75
& 53.33 / 63.33 \\
\midrule
\textbf{Action$\rightarrow$Hybrid}
& 63.86 / 79.52
& \textbf{50.20 / 67.97}
& \textbf{70.00 / 77.50}
& \textbf{57.92 / 66.67} \\
\bottomrule
\end{tabular}
\end{table}

\vspace{-0.5em}
\section{Conclusion}
\vspace{-0.5em}

We presented an adaptive training framework for multi-step GUI agents that
jointly addresses \emph{where to learn} and \emph{how far to roll out}.
The framework combines action-disagreement pivot selection, reusable
intermediate checkpoints, and adaptive Local, Suffix, and Root training.
Across MobileGym and CUA-Gym-Hub, it improves training efficiency by reducing
both optimization samples and environment interaction while maintaining or
improving task performance. These results show that selectively allocating
training across intermediate states and rollout ranges is an effective way to
make GUI-agent RL more efficient.

\newpage
\bibliography{iclr2027_conference}

@misc{qwenuiagent,
  title={Qwen-UI-Agent Technical Report: Toward Next-Generation Real-World Centric Foundation GUI Agents}, 
      author={Hanzhang Zhou and Panrong Tong and Xu Zhang and Quyu Kong and Chenglin Cai and Tianyu Xia and Gongjie Zhang and Jianan Zhang and Long Li and Long Chen and Lei Wang and Gaole Dai and Pengxiang Li and Liangyu Chen and Yue Wang and Steven Hoi},
      year={2026},
      eprint={2607.28227},
      archivePrefix={arXiv},
      primaryClass={cs.AI},
      url={https://arxiv.org/abs/2607.28227}, 
}

@misc{mobileguirl,
  title={MobileGUI-RL: Advancing Mobile GUI Agent through Reinforcement Learning in Online Environment}, 
      author={Yucheng Shi and Wenhao Yu and Zaitang Li and Yonglin Wang and Hongming Zhang and Ninghao Liu and Haitao Mi and Dong Yu},
      year={2025},
      eprint={2507.05720},
      archivePrefix={arXiv},
      primaryClass={cs.LG},
      url={https://arxiv.org/abs/2507.05720}, 
}

@misc{mobilerl,
  title={MobileRL: Online Agentic Reinforcement Learning for Mobile GUI Agents}, 
      author={Yifan Xu and Xiao Liu and Xinghan Liu and Jiaqi Fu and Hanchen Zhang and Bohao Jing and Shudan Zhang and Yuting Wang and Wenyi Zhao and Yuxiao Dong},
      year={2025},
      eprint={2509.18119},
      archivePrefix={arXiv},
      primaryClass={cs.LG},
      url={https://arxiv.org/abs/2509.18119}, 
}

@misc{dartgui,
  title={Efficient Multi-turn RL for GUI Agents via Decoupled Training and Adaptive Data Curation}, 
  author={Pengxiang Li and Zechen Hu and Zirui Shang and Jingrong Wu and Yang Liu and Hui Liu and Zhi Gao and Chenrui Shi and Bofei Zhang and Zihao Zhang and Xiaochuan Shi and Zedong YU and Yuwei Wu and Xinxiao Wu and Yunde Jia and Liuyu Xiang and Zhaofeng He and Qing Li},
  year={2025},
  eprint={2509.23866},
  archivePrefix={arXiv},
  primaryClass={cs.LG},
  url={https://arxiv.org/abs/2509.23866}, 
}

@misc{cuagym,
      title={CUA-Gym: Scaling Verifiable Training Environments and Tasks for Computer-Use Agents}, 
      author={Bowen Wang and Dunjie Lu and Junli Wang and Tianyi Bai and Shixuan Liu and Zhipeng Zhang and Haiquan Wang and Hao Hu and Tianbao Xie and Shuai Bai and Dayiheng Liu and Que Shen and Junyang Lin and Tao Yu},
      year={2026},
      eprint={2605.25624},
      archivePrefix={arXiv},
      primaryClass={cs.AI},
      url={https://arxiv.org/abs/2605.25624}, 
}

@misc{guir1,
  title={GUI-R1 : A Generalist R1-Style Vision-Language Action Model For GUI Agents}, 
      author={Run Luo and Lu Wang and Wanwei He and Longze Chen and Jiaming Li and Xiaobo Xia},
      year={2025},
      eprint={2504.10458},
      archivePrefix={arXiv},
      primaryClass={cs.CV},
      url={https://arxiv.org/abs/2504.10458}, 
}

@misc{uitars,
      title={UI-TARS: Pioneering Automated GUI Interaction with Native Agents}, 
      author={Yujia Qin and Yining Ye and Junjie Fang and Haoming Wang and Shihao Liang and Shizuo Tian and Junda Zhang and Jiahao Li and Yunxin Li and Shijue Huang and Wanjun Zhong and Kuanye Li and Jiale Yang and Yu Miao and Woyu Lin and Longxiang Liu and Xu Jiang and Qianli Ma and Jingyu Li and Xiaojun Xiao and Kai Cai and Chuang Li and Yaowei Zheng and Chaolin Jin and Chen Li and Xiao Zhou and Minchao Wang and Haoli Chen and Zhaojian Li and Haihua Yang and Haifeng Liu and Feng Lin and Tao Peng and Xin Liu and Guang Shi},
      year={2025},
      eprint={2501.12326},
      archivePrefix={arXiv},
      primaryClass={cs.AI},
      url={https://arxiv.org/abs/2501.12326}, 
}

@misc{uir1,
      title={UI-R1: Enhancing Efficient Action Prediction of GUI Agents by Reinforcement Learning}, 
      author={Zhengxi Lu and Yuxiang Chai and Yaxuan Guo and Xi Yin and Liang Liu and Hao Wang and Han Xiao and Shuai Ren and Guanjing Xiong and Hongsheng Li},
      year={2025},
      eprint={2503.21620},
      archivePrefix={arXiv},
      primaryClass={cs.AI},
      url={https://arxiv.org/abs/2503.21620}, 
}

@misc{showui,
      title={ShowUI: One Vision-Language-Action Model for GUI Visual Agent}, 
      author={Kevin Qinghong Lin and Linjie Li and Difei Gao and Zhengyuan Yang and Shiwei Wu and Zechen Bai and Weixian Lei and Lijuan Wang and Mike Zheng Shou},
      year={2024},
      eprint={2411.17465},
      archivePrefix={arXiv},
      primaryClass={cs.CV},
      url={https://arxiv.org/abs/2411.17465}, 
}

@misc{treegrpo,
  title={Tree Search for LLM Agent Reinforcement Learning}, 
      author={Yuxiang Ji and Ziyu Ma and Yong Wang and Guanhua Chen and Xiangxiang Chu and Liaoni Wu},
      year={2026},
      eprint={2509.21240},
      archivePrefix={arXiv},
      primaryClass={cs.LG},
      url={https://arxiv.org/abs/2509.21240}, 
}

@misc{osexpert,
  title={OSExpert: Computer-Use Agents Learning Professional Skills via Exploration}, 
      author={Jiateng Liu and Zhenhailong Wang and Rushi Wang and Bingxuan Li and Jeonghwan Kim and Aditi Tiwari and Pengfei Yu and Denghui Zhang and Heng Ji},
      year={2026},
      eprint={2603.07978},
      archivePrefix={arXiv},
      primaryClass={cs.AI},
      url={https://arxiv.org/abs/2603.07978}, 
}

@misc{mobilegym,
  title={MobileGym: A Verifiable and Highly Parallel Simulation Platform for Mobile GUI Agent Research}, 
      author={Dingbang Wu and Rui Hao and Haiyang Wang and Shuzhe Wu and Han Xiao and Zhenghong Li and Bojiang Zhou and Zheng Ju and Zichen Liu and Lue Fan and Zhaoxiang Zhang},
      year={2026},
      eprint={2605.26114},
      archivePrefix={arXiv},
      primaryClass={cs.AI},
      url={https://arxiv.org/abs/2605.26114}, 
}

@misc{pivotrl,
  title={PivotRL: High Accuracy Agentic Post-Training at Low Compute Cost}, 
      author={Junkeun Yi and Damon Mosk-Aoyama and Baihe Huang and Ritu Gala and Charles Wang and Sugam Dipak Devare and Khushi Bhardwaj and Abhibha Gupta and Oleksii Kuchaiev and Jiantao Jiao and Jian Zhang and Venkat Srinivasan},
      year={2026},
      eprint={2603.21383},
      archivePrefix={arXiv},
      primaryClass={cs.AI},
      url={https://arxiv.org/abs/2603.21383}, 
}

@misc{pivoarl,
  title={Agent Reinforcement Learning via Pivotal-Aware Self-Feedback Retry}, 
      author={Weiyang Guo and Zesheng Shi and Longhui Zhang and Zeen Zhu and Min Zhang and Jing Li},
      year={2026},
      eprint={2607.03702},
      archivePrefix={arXiv},
      primaryClass={cs.AI},
      url={https://arxiv.org/abs/2607.03702}, 
}

@misc{bpo,
  title={Branching Policy Optimization: Sandbox-Native Language Agent Reinforcement Learning}, 
      author={Bowei He and Yankai Chen and Xiaokun Zhang and Xue Liu},
      year={2026},
      eprint={2607.14171},
      archivePrefix={arXiv},
      primaryClass={cs.LG},
      url={https://arxiv.org/abs/2607.14171}, 
}

@misc{src,
  title={Speculative Rollback Correction for Quality-Diverse Web Agent Imitation}, 
      author={Longkun Hao and Hongyu Lin and Hao Li and Zhuowen Liu and Zhichao Yang and Haojie Hao and Dongshuo Huang and Haitao Yang and Hongyu Ge and Ming jie Xie and Yanjun Wu and Zi Hao Yin and Yan Bai and Yihang Lou},
      year={2026},
      eprint={2606.12485},
      archivePrefix={arXiv},
      primaryClass={cs.LG},
      url={https://arxiv.org/abs/2606.12485}, 
}
\bibliographystyle{iclr2027_conference}

\appendix
\section{Appendix}

\subsection{Environment Checkpointing}

We use Podman and CRIU to checkpoint and restore interactive environments at the container level. CRIU captures running process state, including browser memory and the JavaScript heap, while Podman packages the checkpoint with container filesystem changes. This preserves runtime information beyond what screenshots or browser storage alone can capture.

Checkpoints are created between agent actions, and the source container can continue running after capture. To resume execution, the saved checkpoint is restored into a target container, together with the corresponding execution metadata and agent context. A single checkpoint can initialize multiple independent rollout branches, allowing exploration from the same intermediate state without repeating earlier actions. Existing TCP connections are closed during restoration so that each branch can establish fresh connections.

We check restoration readiness through backend health checks and successful screenshot capture. Additional consistency tests compare observations and task-evaluation results before and after restoration. These checks help identify observable restoration failures, but do not constitute proof of complete state equivalence or guarantee identical future execution when asynchronous events or external services are involved.

\subsection{Action Disagreement and Local RL}
\label{app:action-matching}

\paragraph{Action matching.}
Both action-disagreement estimation and local one-step RL require determining
whether two GUI actions should be treated as equivalent. For coordinate-based
actions, including click, long-press, and double-click, two actions are
considered equivalent if their coordinates differ by at most $5\%$ of the
screen width and height. Swipe actions are matched according to their
discretized direction. For other actions, we compare the action type directly.
In particular, text-entry actions are treated as the same \texttt{type} action
regardless of the generated text content. This rule removes minor
coordinate or textual variation that does not correspond to a different GUI
decision.

\paragraph{Action disagreement.}
Given $N=8$ policy samples, we first group sampled responses according to the
action-matching rule above and compute $\hat{D}(c_t)$ from the resulting action
counts. We use $\tau_{\mathrm{dis}}=0.45$ in our experiments. For example, a
$6$--$2$ split yields $\hat{D}=0.429$ and is filtered out, whereas
$6$--$1$--$1$ ($\hat{D}=0.464$) and $5$--$3$ ($\hat{D}=0.536$) are retained. Detailed disagreements are listed in Table~\ref{tab:disagreement-patterns}.

\paragraph{Local one-step reward.}
The same action-matching rule is used to construct the local reward. Let
$a_t^{E}$ denote the expert next action at pivot $c_t$, and let
$a_t^{(i)}$ be a policy sample. We assign

$$
r_t^{(i)}=\mathbb{I}\!\left[\mathcal{M}(a_t^{(i)},a_t^{E})\right],
$$

where $\mathcal{M}$ is the action-matching function defined above. The expert
prefix is used only to condition the policy at the pivot; optimization is
performed on the newly sampled response.

\begin{table}[t]
\centering
\small
\caption{
Representative action-count patterns and their empirical disagreement scores
for $N=8$ policy samples. We use $\tau_{\mathrm{dis}}=0.45$.
}
\label{tab:disagreement-patterns}
\setlength{\tabcolsep}{10pt}
\renewcommand{\arraystretch}{1.08}
\begin{tabular}{cc@{\hspace{12mm}}cc}
\toprule
\textbf{Action Counts} & $\hat{D}(c_t)$ &
\textbf{Action Counts} & $\hat{D}(c_t)$ \\
\midrule
$8$               & $0.000$ &
$5$--$3$          & $0.536$ \\

$7$--$1$          & $0.250$ &
$4$--$4$          & $0.571$ \\

$6$--$2$          & $0.429$ &
$5$--$2$--$1$     & $0.607$ \\

$6$--$1$--$1$     & $0.464$ &
$4$--$3$--$1$     & $0.679$ \\

$4$--$2$--$2$     & $0.714$ &
$3$--$3$--$2$     & $0.750$ \\

$2$--$2$--$2$--$2$ & $0.857$ &
$1$--$1$--$1$--$1$--$1$--$1$--$1$--$1$ & $1.000$ \\
\bottomrule
\end{tabular}
\end{table}

\subsection{Adaptive Hybrid Training Algorithm}

Pivots are ordered according to their positions in the expert trajectory, with
$c_K$ denoting the deepest pivot. Hybrid training is monotonic with respect to
the rollout frontier: Suffix training starts from $c_K$ and moves one pivot
earlier whenever the current suffix success rate exceeds
$\tau_{\mathrm{suffix}}$. Once the frontier reaches the task root, the task
remains in the Root regime and is not routed back to intermediate pivots.
In our experiments, we set
$\tau_{\mathrm{low}}=0.25$,
$\tau_{\mathrm{high}}=0.75$, and
$\tau_{\mathrm{suffix}}=0.5$. Algorithm~\ref{alg:hybrid} shows the entire pipeline.

\begin{algorithm}[t]
\caption{Adaptive Hybrid Training}
\label{alg:hybrid}
\begin{algorithmic}[1]
\REQUIRE Task $x$, ordered pivots
$\mathcal{C}(x)=\{c_1,\ldots,c_K\}$ from earliest to deepest,
policy $\pi$, thresholds
$\tau_{\mathrm{low}}$, $\tau_{\mathrm{high}}$,
and $\tau_{\mathrm{suffix}}$
\ENSURE Updated policy $\pi$

\STATE Run $4$ root-level rollouts and compute success rate $\hat{p}(x)$
\IF{$\hat{p}(x)<\tau_{\mathrm{low}}$}
    \STATE $R \leftarrow \textsc{Local}$
\ELSIF{$\hat{p}(x)\leq\tau_{\mathrm{high}}$}
    \STATE $R \leftarrow \textsc{Suffix}$; $j \leftarrow K$
\ELSE
    \STATE $R \leftarrow \textsc{Root}$
\ENDIF

\WHILE{training continues}
    \IF{$R=\textsc{Local}$}
        \STATE Perform one-step RL at selected pivots using $8$ policy samples
        \STATE Run $4$ root-level rollouts and recompute $\hat{p}(x)$
        \IF{$\hat{p}(x)<\tau_{\mathrm{low}}$}
            \STATE $R \leftarrow \textsc{Local}$
        \ELSIF{$\hat{p}(x)\leq\tau_{\mathrm{high}}$}
            \STATE $R \leftarrow \textsc{Suffix}$; $j \leftarrow K$
        \ELSE
            \STATE $R \leftarrow \textsc{Root}$
        \ENDIF

    \ELSIF{$R=\textsc{Suffix}$}
        \STATE Restore pivot $c_j$ and generate $8$ suffix rollouts
        \STATE Update $\pi$ using the resulting rollout group
        \STATE Compute suffix success rate $\hat{p}_{\mathrm{suf}}$
        \IF{$\hat{p}_{\mathrm{suf}}>\tau_{\mathrm{suffix}}$}
            \IF{$j>1$}
                \STATE $j \leftarrow j-1$
            \ELSE
                \STATE $R \leftarrow \textsc{Root}$
            \ENDIF
        \ENDIF

    \ELSE
        \STATE Generate $2$ root-level rollouts
        \IF{the rollout group exhibits outcome variation}
            \STATE Update $\pi$ using the root rollout group
        \ENDIF
        \STATE Keep $R \leftarrow \textsc{Root}$
    \ENDIF
\ENDWHILE
\end{algorithmic}
\end{algorithm}

\subsection{AI Assistance Statement}

We used generative AI tools to assist with writing and language polishing,
literature retrieval and discovery, and the generation of synthetic data used
in our research workflow. AI-assisted text was reviewed and revised by the
authors, and retrieved literature was manually verified against the original
sources. Any AI-generated data used in experiments was inspected and validated
before use. The authors take full responsibility for the final content,
experimental results, and claims in this paper.

\end{document}